\documentclass[letterpaper, 10 pt, conference]{ieeeconf}  

\IEEEoverridecommandlockouts                              

\usepackage[hidelinks]{hyperref}

\usepackage{graphicx}
\usepackage{float}
\usepackage{amsmath} 

\title{\LARGE \bf
Leveraging VR Robot Games to Facilitate Data Collection for Embodied Intelligence Tasks
}

\author{
\authorblockN{
Yihan Zhang$^{1*}$, 
Ziyun Huang$^{1*}$, 
and Linqi Ye$^{2}$
}
\thanks{*These authors contributed equally to this work.}
\thanks{$^{1}$Y. Zhang and Z. Huang are with the School of Computer Engineering and Science, Shanghai University, Shanghai, 200444, China
}
\thanks{$^{2}$L. Ye is with the School of Future Technology, Shanghai University, Shanghai, 200444, China (Corresponding author, email: yelinqi@shu.edu.cn)}
}

\begin{document}

\maketitle
\thispagestyle{empty}
\pagestyle{empty}

\begin{abstract}
Collecting embodied interaction data at scale remains costly and difficult due to the  limited accessibility of conventional interfaces. We present a gamified data collection framework based on Unity that combines procedural scene generation, VR-based humanoid robot control, automatic task evaluation, and trajectory logging. A trash pick-and-place task prototype is developed to validate the full workflow.Experimental results indicate that the collected demonstrations exhibit broad coverage of the state-action space, and that increasing task difficulty leads to higher motion intensity as well as more extensive exploration of the arm's workspace. The proposed framework demonstrates that game-oriented virtual environments can serve as an effective and extensible solution for embodied data collection.
\end{abstract}

\section{INTRODUCTION}
Embodied intelligence demands large-scale interaction data collected across diverse environments and tasks for effective training. However, existing data collection pipelines rely heavily on expert demonstrations, manually built simulation scenarios, or costly real-world robotic operations. Consequently, scaling both the quantity and diversity of data at low cost remains a substantial challenge. In addition, many traditional robot control or teleoperation interfaces require sufficient training, which limits the accessibility and increases deployment cost.

In this work, we explore a game-oriented approach to embodied data collection in Unity. Our key idea embeds embodied task execution into an interactive virtual environment, generating usable task data through gameplay. To boost accessibility, our system uses VR-based control, offering a more intuitive interface than traditional robot operation and lowering barriers for non-expert users. This design has the potential to reduce personnel training overhead and interface migration cost while making data collection more engaging and easier to scale.

Our workflow integrates procedural scene generation, task-oriented interaction design, automatic task completion evaluation, and structured data gathering. Procedural generation improves scene diversity with minimal manual effort, while the evaluation and logging modules transform each interaction episode into embodied task data, such as motion trajectories and completion time. In addition, Unity-based data collection introduces greater variability (e.g., lighting and environment), provides a exhausive component library, and enables more robust error handling. The framework is also extensible to more complex interactive scenarios, including competitive and multi-agent settings. In addition, Unity-based data collection provides higher randomness (e.g., lighting and environmental variations), offers a rich component library, and supports more robust error handling. The framework is designed with extensibility towards richer interactive settings, including competitive and multi-agent gameplay.

We implement a demo task where a robot picks up rubbish and places it into a bin. The demo supports randomized scene generation, automatic task completion evaluation, VR-based interaction, and motion trajectory and completion time logging.  Although still a prototype, our system establishes an end-to-end workflow for converting game-like embodied interaction into structured data. The following are our contributions:

(1) We propose a gamified framework for embodied data collection that reframes data acquisition as an interactive gameplay process.

(2) We develop a Unity-based prototype integrating procedural environment generation, VR-based control, automatic task assessment, and trajectory logging.

(3) We demonstrate the feasibility of this framework as a scalable and extensible direction for collecting embodied interaction data with lower participation barriers.

\begin{figure}[H]
  \centering
  \includegraphics[width=0.48\textwidth]{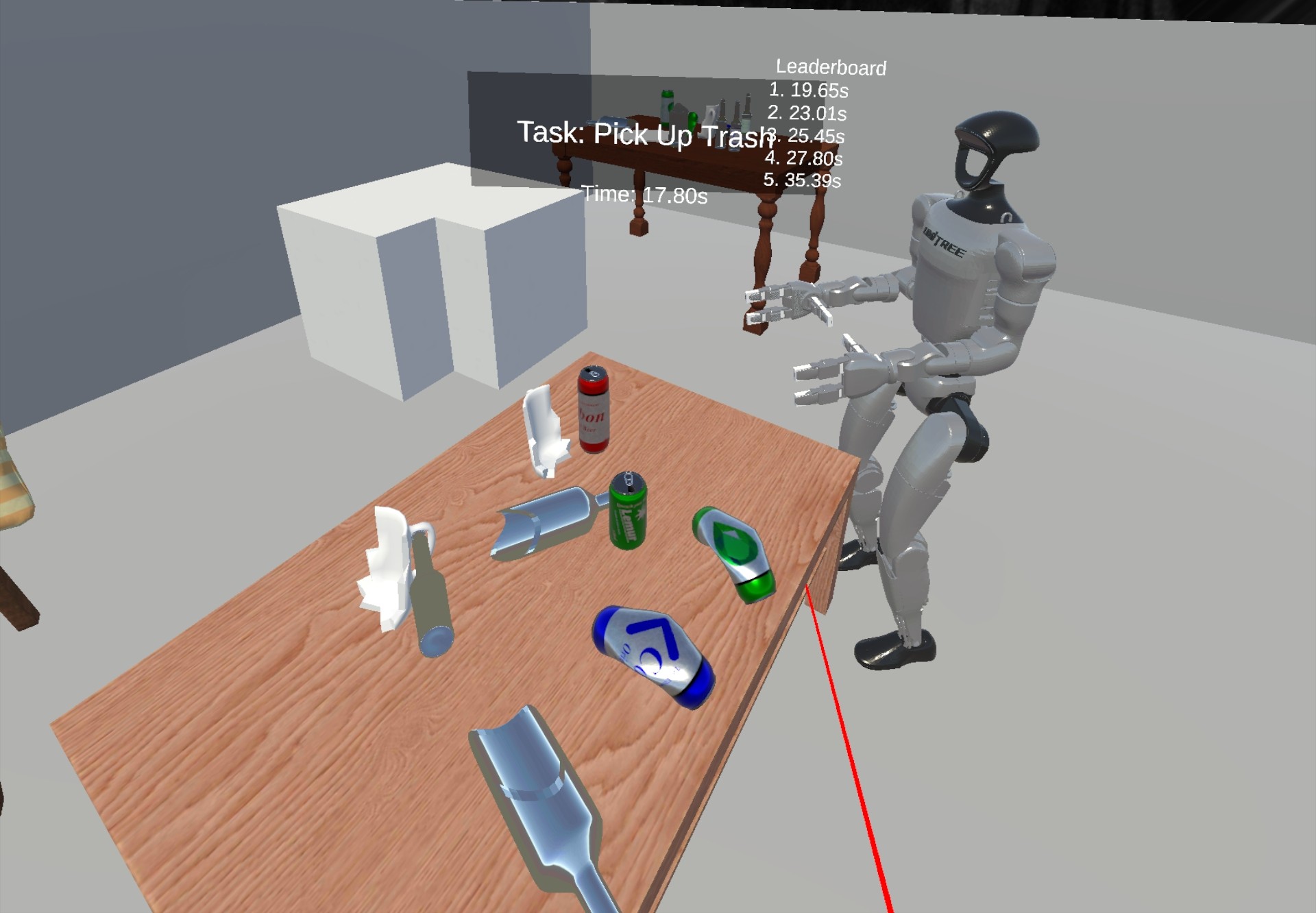}
  \caption{Screenshot of our prototype.}
  \label{fig:control}
\end{figure}

\section{RELATED WORK}
\subsection{Embodied Data Collection and Learning from Demonstrations}

Data collection remains a major bottleneck in embodied intelligence. A large body of work studies robot learning from demonstration (LfD), where policies are learned from human-provided trajectories or expert demonstrations. As summarized by Argall et al.~\cite{argall2009survey}, demonstration-based pipelines are effective for grounding robot behaviors, but they often rely on expert operators, carefully controlled setups, and labor-intensive collection procedures. While such methods can provide high-quality supervision, they are expensive to scale and often limited in environmental and behavioral diversity.

To address scalability, recent work has explored distributed and large-scale data collection pipelines. RoboTurk~\cite{mandlekar2019roboturk} shows that robotic manipulation data can be collected from many remote users, substantially expanding supervision beyond traditional lab-based collection. More recently, Open X-Embodiment~\cite{openx2024} demonstrates the value of aggregating heterogeneous robot datasets across embodiments, tasks, and platforms. These works highlight the importance of scale and diversity in embodied data, but they primarily focus on building larger datasets through specialized collection interfaces rather than designing the collection process itself to be broadly engaging.

Another related direction studies less structured forms of human interaction data. Learning Latent Plans from Play~\cite{lynch2019play} argues that play data can be easier to collect than narrowly defined task demonstrations because it does not require precise task segmentation or repeated manual resets. MimicPlay~\cite{wang2023mimicplay} further shows that human play can provide rich supervision for long-horizon imitation learning. These works motivate our setting by suggesting that open-ended interaction can produce valuable embodied data at lower annotation cost. However, they do not explicitly frame data collection as a gamified process with designed incentives, procedural task variation, and built-in task evaluation. Our work builds on this intuition and investigates how game mechanics can further broaden participation and support sustained embodied data collection.

\subsection{Embodied Simulators and Procedurally Generated Environments}

Simulation platforms have become foundational infrastructure for embodied AI research. Interactive 3D environments such as AI2-THOR~\cite{kolve2017ai2thor}, Habitat~\cite{savva2019habitat}, and RoboTHOR~\cite{deitke2020robothor} provide controllable, scalable, and reproducible settings for embodied agents to perceive, navigate, and interact with objects. These simulators have been widely used for training and benchmarking embodied policies, especially when real-world data collection is expensive or difficult to standardize.

Beyond fixed environments, procedural generation has emerged as an important strategy for improving scene diversity and coverage. ProcTHOR~\cite{deitke2022procthor} demonstrates that large numbers of interactive indoor environments can be generated procedurally, enabling broader scene distributions than hand-authored benchmarks alone. More generally, procedural content generation (PCG) has been extensively studied in games as a way to reduce manual authoring effort while increasing replayability and diversity~\cite{hendrikx2013pcg}. Prior work has also explored experience-driven and learning-based PCG, where generated content is shaped not only by randomness but also by player experience or optimization objectives~\cite{yannakakis2011edpcg,khalifa2020pcgrl}.

These works directly motivate our use of random scene generation. However, most embodied simulators and procedurally generated environments are primarily designed for agent training, benchmarking, or simulation-to-real transfer. In contrast, our focus is on using procedurally generated Unity environments as a human-facing data collection layer, where diversity is valuable not only for improving coverage but also for sustaining replayability and enlarging the behavioral space induced by player interaction.

\subsection{Positioning of Our Work}

Our work lies at the intersection of embodied data collection, interactive simulation, procedural environment generation, and gamified participation. Prior work has established the value of large-scale robot datasets, play-based interaction data, and embodied simulators as separate research directions. However, these directions are often pursued in isolation: embodied AI platforms mainly support training and evaluation, while gamification is rarely applied to embodied tasks. In contrast, we propose a unified framework in which game mechanics serve as the incentive layer, procedural scene generation serves as the diversity layer, and robot-oriented task interfaces serve as the embodied interaction layer. This combination allows data collection to become more scalable, replayable, and extensible.
\section{METHODOLOGY}
\begin{figure}[H]
\centering
\includegraphics[width=0.48\textwidth,height=4.5cm,keepaspectratio]{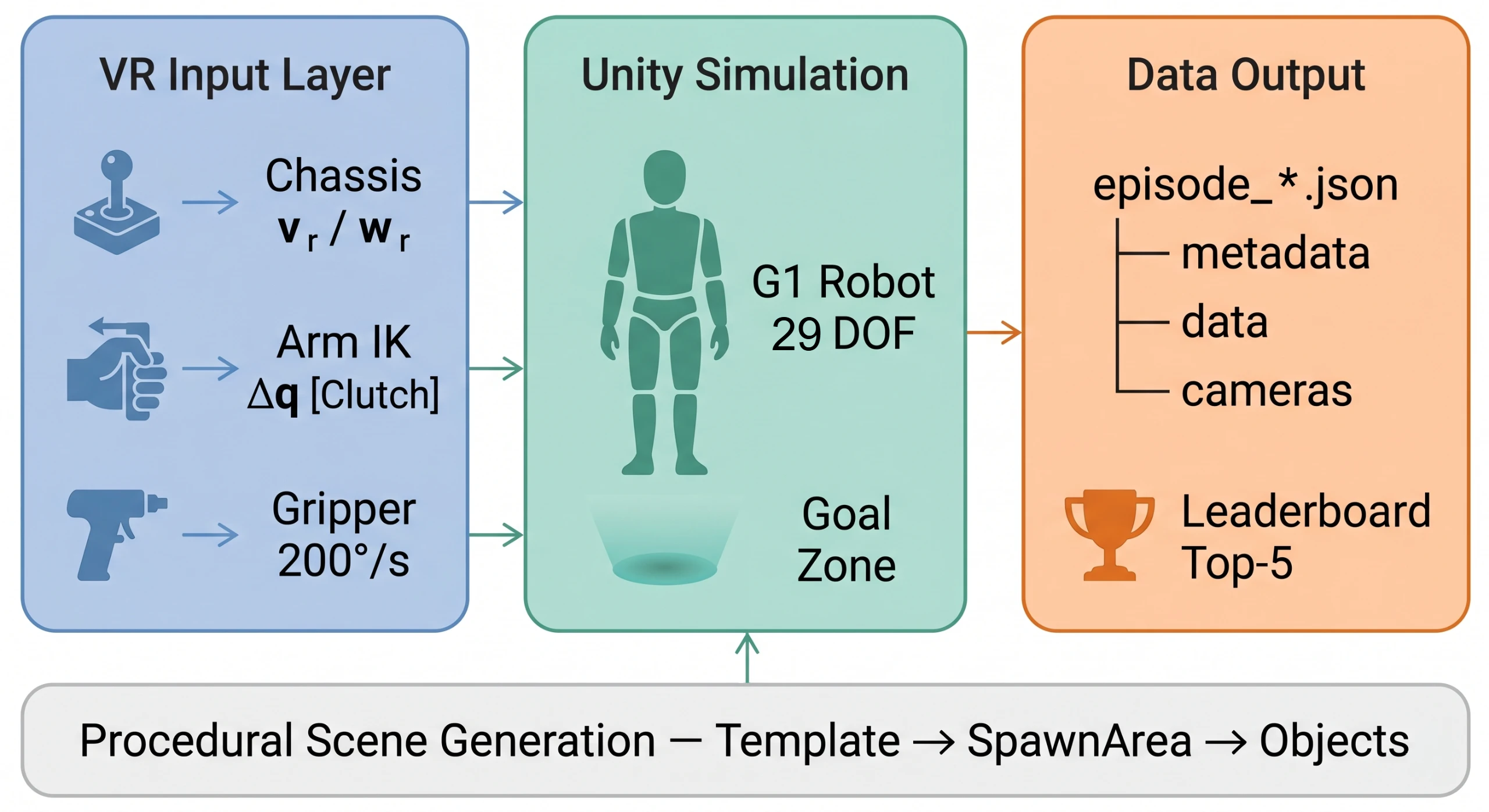}
\caption{Overall system architecture. Procedural scene generation populates a Unity environment containing a Unitree G1 humanoid robot. The user controls the robot via a PICO Neo3 headset; raw controller inputs are translated by the VR Control Adapter into chassis velocity, arm IK targets, and gripper commands. Successful episodes trigger trajectory saving and leaderboard updates.}
\label{fig:arch}
\end{figure}

The system consists of four interconnected modules: procedural scene generation, VR interaction and motion mapping, task management and completion assessment, and data collection. Fig.~\ref{fig:arch} illustrates the overall architecture.

\subsection{Procedural Scene Generation}
\begin{figure}[H]
  \centering
  \includegraphics[width=0.48\textwidth]{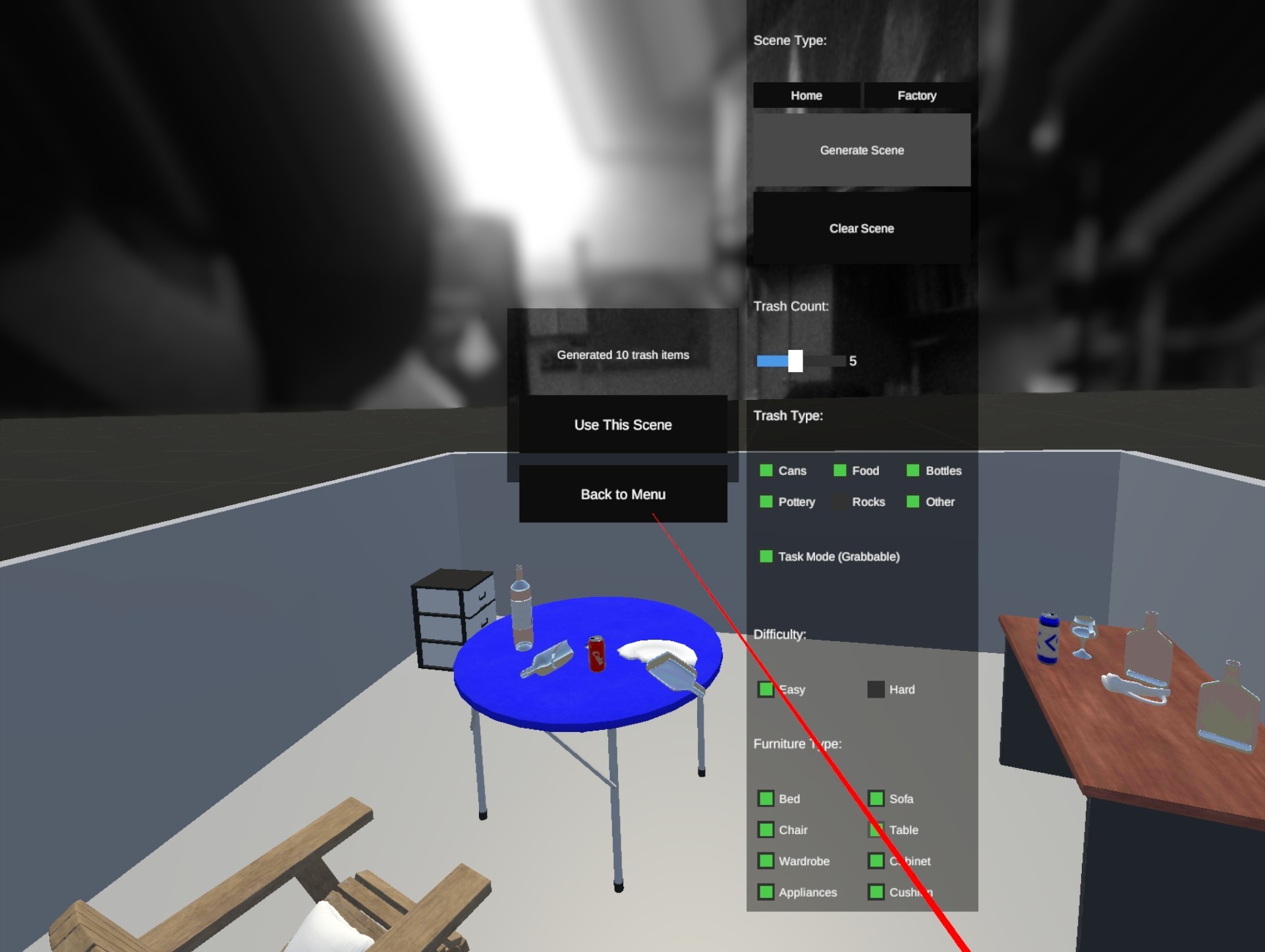}
  \caption{A screenshot of the generated scene}
  \label{fig:generated_scene}
\end{figure}

To generate diverse indoor task environments at low manual cost, we adopt a simple rule-based method to place objects automatically. a room layout is picked at random from a set of ready-made templates each of which contains \textit{SpawnArea} regions that specify allowed object categories (\textit{objectType}) and a maximum object count (\textit{maxObjects}). Within these constraints, the scene is populated in stages: trash bins $\rightarrow$ tables $\rightarrow$ furniture $\rightarrow$ decorations $\rightarrow$ trash (task objects). Each placement candidate is validated by downward ray-casting to determine height, overlap detection to avoid collisions, and boundary checking to stay within the spawn region. Only physically feasible placements are accepted; in task mode, trash objects are tagged as interactable.

\begin{figure}[H]
\centering
\includegraphics[width=0.48\textwidth,keepaspectratio,height=6cm]{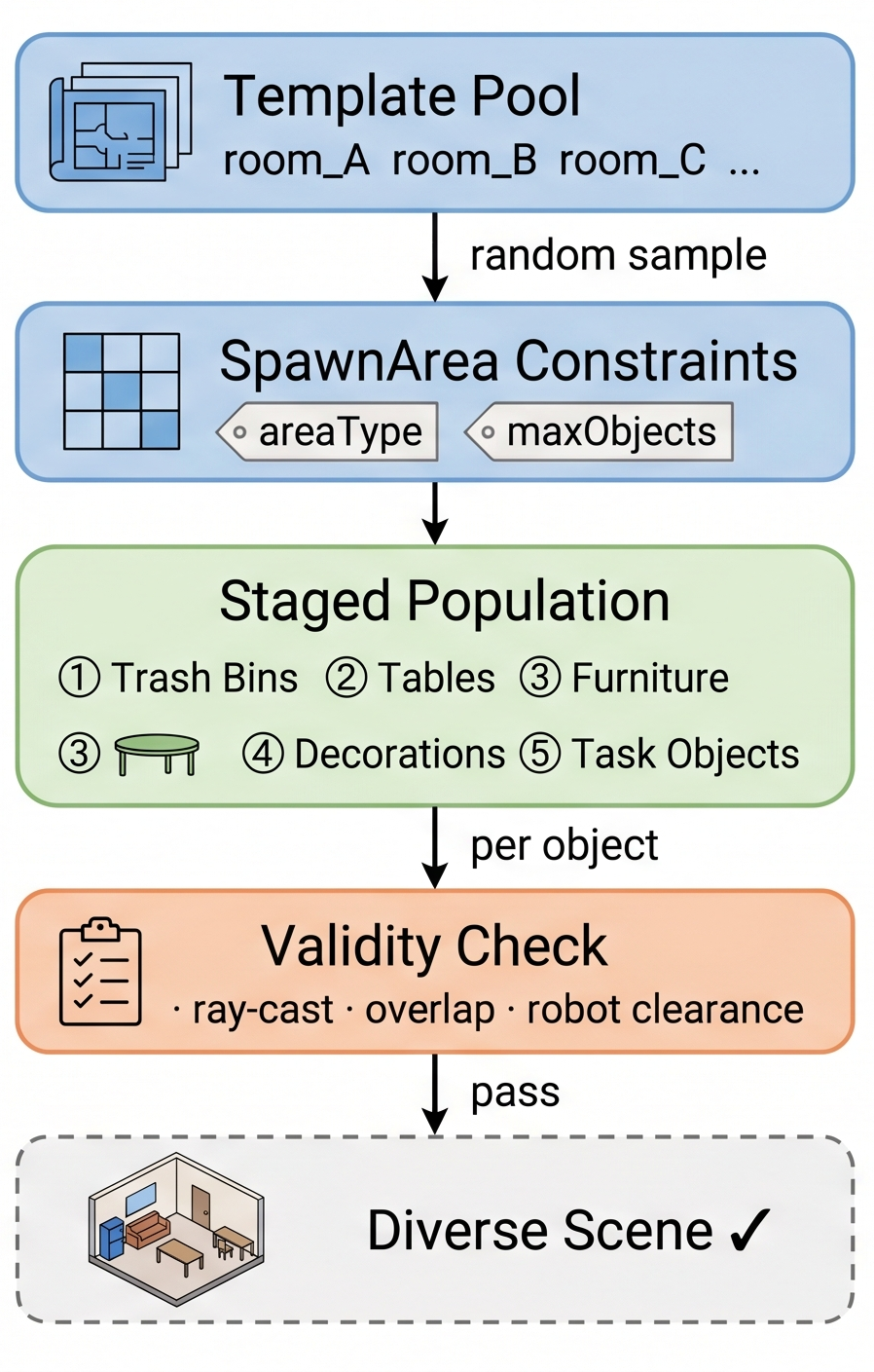}
\caption{Procedural scene generation pipeline. A room template is selected, \textit{SpawnArea} constraints are read, and objects are placed in a fixed order with validity filtering at each stage.}
\label{fig:gen}
\end{figure}

\subsection{VR Interaction and Motion Mapping}
A PICO Neo3 headset with dual controllers provides the human-robot interface. The \textit{VRControlAdapter} module maps raw inputs to robot commands across three layers.

\textbf{Chassis control.} A reinforcement learning–pretrained neural network is employed to control the robot’s lower limbs for locomotion~\cite{ye2026gewu}. The left thumbstick axes are mapped to chassis forward velocity command $v_r$ and rotational angular velocity command $w_r$.

\textbf{Arm IK control (Clutch mode).} To prevent abrupt arm jumps, IK targets are updated only while the Grip button is held. On first press, the controller world-space position $\mathbf{p}_\text{grip}$ and current IK target $\mathbf{q}_0$ are recorded. During continued holding, the controller displacement $\Delta\mathbf{p} = \mathbf{p}_\text{current} - \mathbf{p}_\text{grip}$ is transformed into the VR camera's local frame and scaled per axis to obtain $\Delta\mathbf{q}$; the IK target is updated as $\mathbf{q} = \mathbf{q}_0 + \Delta\mathbf{q}$, clamped to safety limits ($x\!\in\![-0.08,0.15]$, $y\!\in\![-0.08,0.35]$, $z\!\in\![-0.15,0.08]$\,m for the right hand; symmetric for the left). Wrist rotation is driven by the incremental quaternion of the controller, limited to $\pm 45^\circ$ per axis.

\textbf{Gripper control.} The right Trigger drives six right-hand finger joints; the left Trigger drives six left-hand joints. Joints move toward the target grasp angle at $200^\circ/\mathrm{s}$ and return to open on release.

IK targets are converted by an inverse kinematics module into joint drive commands applied to all 29 DOF via Unity ArticulationBody PD controllers. Body joint commands pass through an exponential smoothing filter:
$$
u_\text{smooth}(t) = \alpha \cdot u_\text{smooth}(t{-}1) + (1-\alpha) \cdot u_\text{cmd}(t), \eqno{(1)}
$$
with $\alpha = 0.9$ to suppress high-frequency jitter.

\begin{figure}[H]
\centering
\includegraphics[width=0.48\textwidth,height=4.5cm,keepaspectratio]{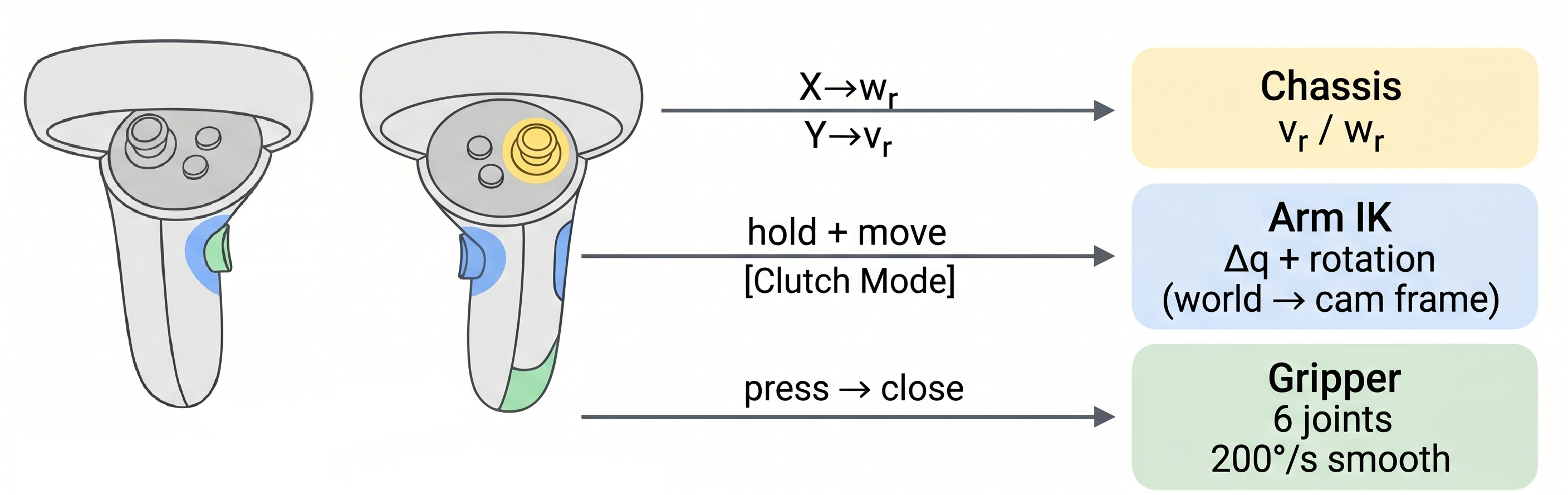}
\caption{VR controller input mapping. Left thumbstick controls chassis motion; Grip activates clutch-mode IK for arm control; Trigger opens/closes the gripper. The right controller follows the same scheme.}
\label{fig:vr}
\end{figure}

\subsection{Task Management and Completion Assessment}
\begin{figure}[H]
\centering
\includegraphics[width=0.48\textwidth,height=4.5cm,keepaspectratio]{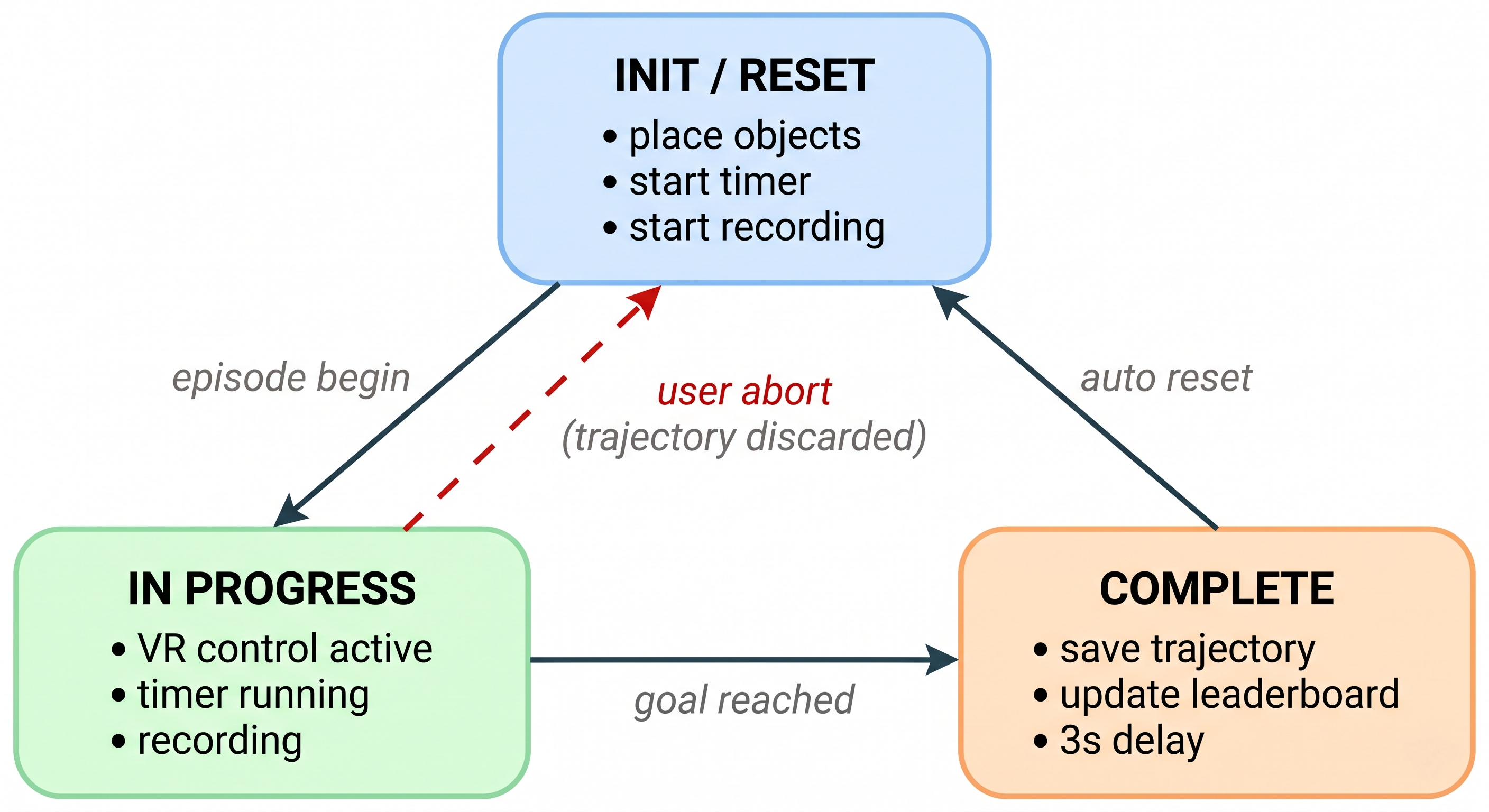}
\caption{Episode state machine. Each episode starts with scene initialization, proceeds through active interaction, and terminates either by successful goal detection or user abort. Successful episodes trigger data saving and leaderboard update before the scene resets.}
\label{fig:fsm}
\end{figure}

Task execution follows a four-phase episode lifecycle: \textit{Init} $\rightarrow$ \textit{InProgress} $\rightarrow$ \textit{Complete/Abort} $\rightarrow$ \textit{Reset}, as shown in Fig.~\ref{fig:fsm}. The goal zone is equipped with a collider; when the target object enters it, the system automatically judges success, records completion time $T_i$, and schedules a reset after a brief delay. No manual annotation is required. Users may abort at any time, in which case the trajectory is discarded. Upon task success, $T_i$ is written to a persistent JSON leaderboard that retains the top-5 records and displays the current rank immediately, motivating players to produce faster and higher-quality demonstrations.

\subsection{Data Collection}
The system records a \textit{FrameData} entry every frame during task execution. Each entry contains: timestamp, positions and velocities of 12 leg joints, 3 waist joints, and 14 arm joints, root node pose, arm IK positions and rotations, and scene object poses. Only successfully completed episodes are saved. Each saved episode is stored as:

\begin{verbatim}
episode_YYYYMMDD_HHmmss/
  metadata.json 
  data.json     (FrameData list)
  cameras/      (optional JPG sequence)
\end{verbatim}

Task-level statistics (completion time $T_i$) are associated with each episode to support behavioral analysis and policy evaluation.

\section{EXPERIMENTS}
This section evaluates the collected embodied interaction data from three perspectives: (1) the basic scale and state--action space coverage quality of the dataset, (2) whether the designed difficulty parameter effectively differentiates human manipulation behavior, and (3) whether the procedural scene generation introduces sufficient environmental diversity. A video demo of the game is available at \url{https://linqi-ye.github.io/video/eai-game.mp4}

\subsection{Dataset Quality Verification}

\subsubsection{Dataset Overview}

Using VR control in the Unity simulation, we collected 17 complete pick-and-place runs under the Easy difficulty, for a total of 26,882 frames, each of which includes a 44-dimensional state (29 body joint positions, 12 hand joint positions, and the 3D position of the robot base) and a 33-dimensional action (chassis speed, arm target positions, gripper commands, and wrist rotation). The data are saved frame by frame in JSON format, along with basic run metadata like total frames and completion time.

\subsubsection{Action Sequence Length Distribution}
Run lengths vary from 28.4 to 58.1 seconds, with an mean of $40.2 \pm 8.7$ s. This spread comes from natural differences in how quickly and smoothly each person controls the arm, which actually adds useful variety to the training data for imitation learning.

\begin{figure}[H]
\centering
\includegraphics[width=0.48\textwidth,height=4.5cm,keepaspectratio]{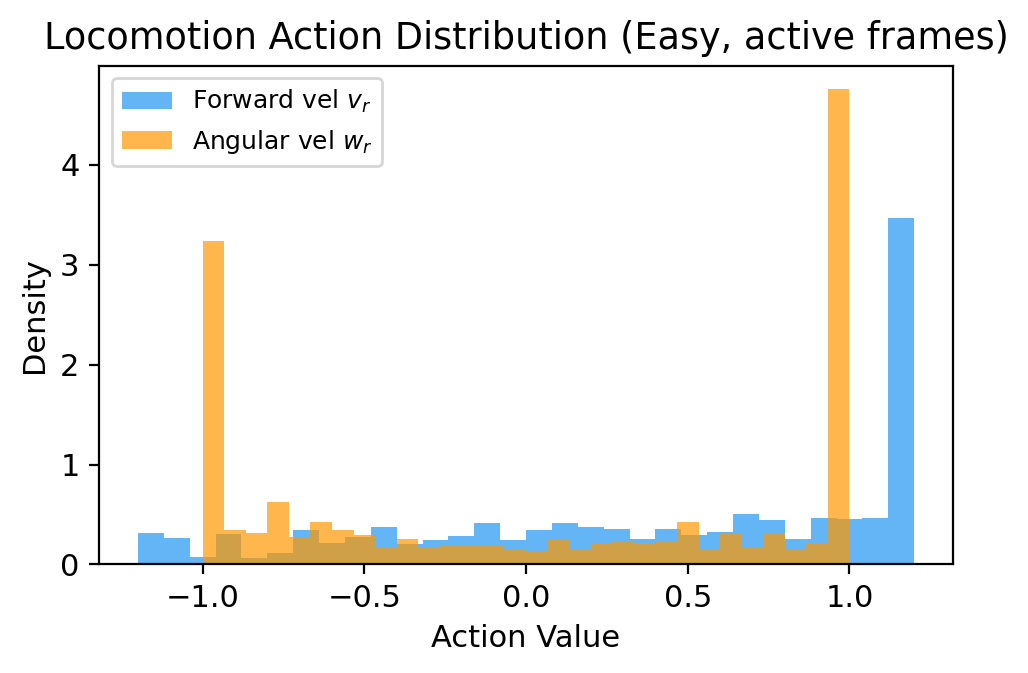}
\caption{Locomotion action distribution (active frames only). Both forward velocity $v_r$ and angular velocity $w_r$ exhibit broad, multimodal distributions, indicating diverse locomotion behaviors across episodes.}
\label{fig:locomotion}
\end{figure}

\subsubsection{State--Action Space Coverage}

To assess the distributional coverage of the collected data, we apply a bin coverage ratio metric: each dimension's value range is divided into 20 equal-width bins, and we compute the fraction of bins containing at least one sample, averaged across dimensions within each subspace. Results are reported in Table~\ref{tab:coverage} and Fig.~\ref{fig:coverage}.

\begin{table}[h]
\centering
\caption{State-Action Space Coverage (Easy, 17 episodes)}
\label{tab:coverage}
\begin{tabular}{lc}
\hline
Subspace & Coverage \\
\hline
Body joints (29-dim)   & 95\% \\
Hand joints (12-dim)   & 95\% \\
Arm IK position (6-dim) & 98\% \\
Chassis motion (2-dim) & 100\% \\
Robot position (3-dim) & 88\% \\
Wrist rotation (12-dim) & 25\% \\
\textbf{Average (excl.\ wrist)} & \textbf{95\%} \\
\hline
\end{tabular}
\end{table}

All major subspaces achieve coverage above 88\%. The low wrist rotation coverage (25\%) is expected: the pick-and-place task requires primarily translational arm motion for grasping, and wrist orientation changes naturally concentrate within a narrow range.

Action activity analysis further shows that chassis motion is active in 46.0\% of frames, arm IK control in 72.5\%, and gripper actuation in 36.4\%, consistent with the typical ``navigate to target, then grasp'' structure of the pick-and-place task.


\begin{figure}[H]
\centering
\includegraphics[width=0.48\textwidth,height=4.5cm,keepaspectratio]{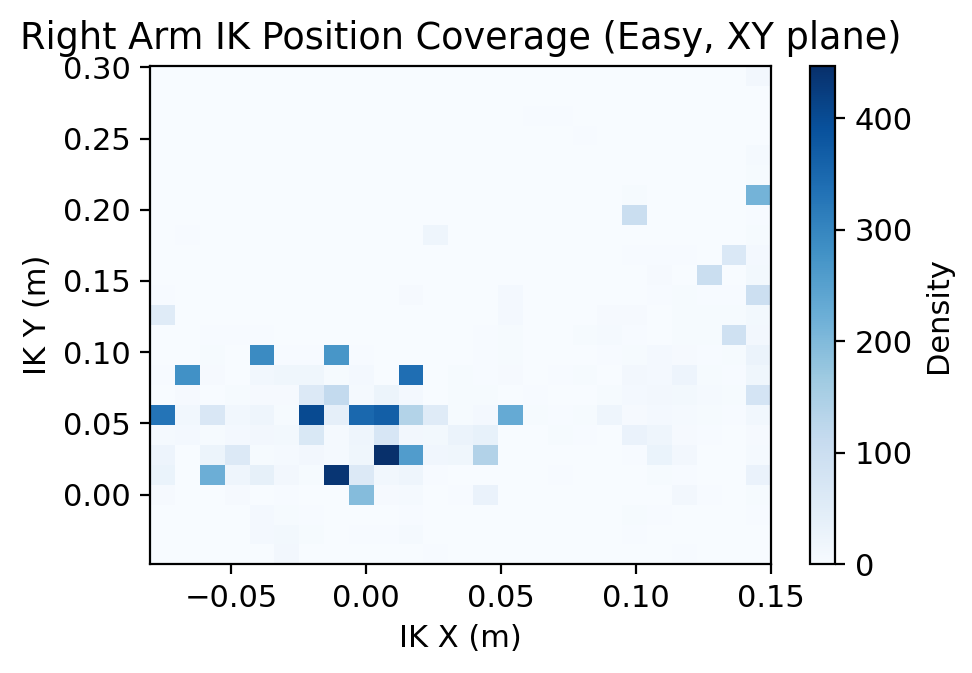}
\caption{Right arm end-effector IK position coverage in the XY plane (active frames only). The distribution spans the reachable workspace, indicating diverse manipulation poses.}
\label{fig:ikheatmap}
\end{figure}

\begin{figure}[H]
\centering
\includegraphics[width=0.48\textwidth,height=4.5cm,keepaspectratio]{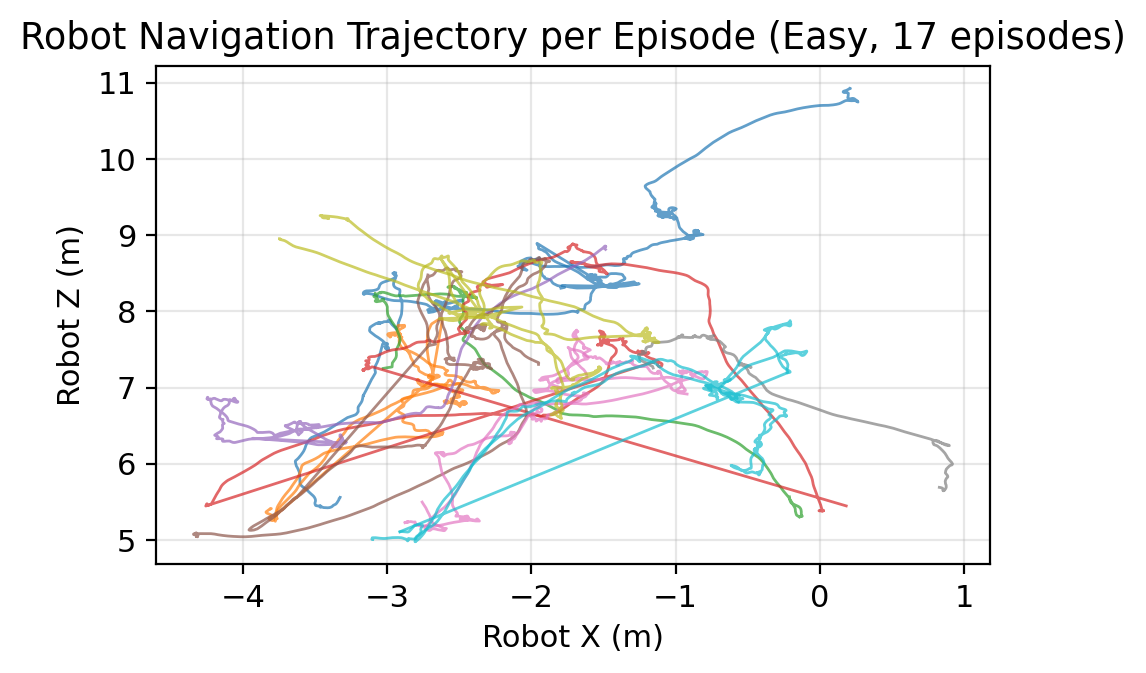}
\caption{Robot navigation trajectories across all 17 episodes. Each colored line represents one episode, showing diverse spatial paths within the procedurally generated scene.}
\label{fig:trajectory}
\end{figure}

Overall, the collected data covers the principal state and action dimensions required for task execution, meeting the basic quality requirements for imitation learning and policy evaluation.

\subsection{Effect of Difficulty Level on the Game}

\subsubsection{Experimental Setup}
\begin{figure}[H]
\centering
\includegraphics[width=0.48\textwidth,height=7cm,keepaspectratio]{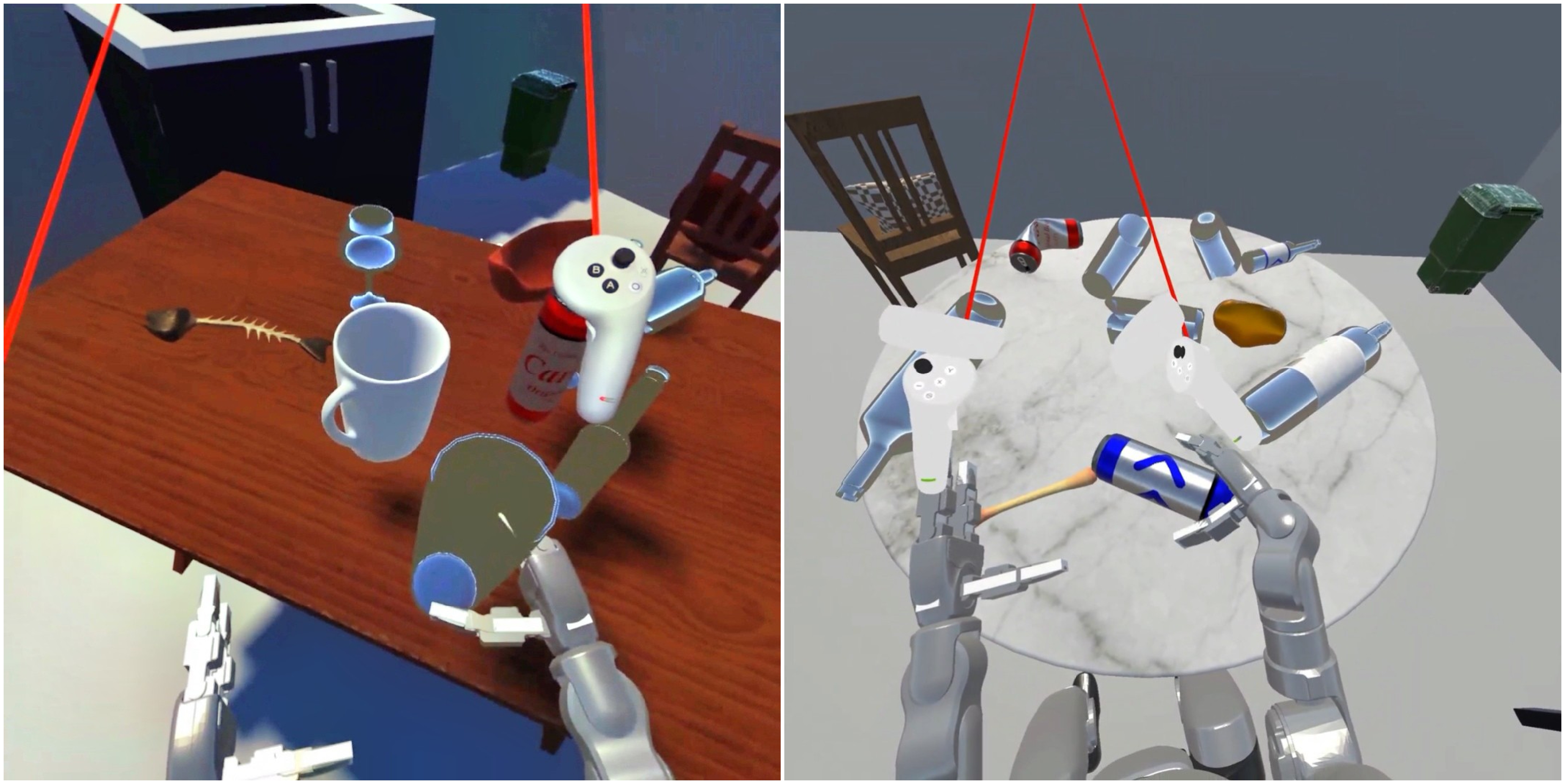}
\caption{Easy vs.\ Hard comparison: game screenshots.}
\label{fig:tasks}
\end{figure}

To validate whether our difficulty setting really changes how people move, we collected data for both Easy and Hard levels using the same basic scene setup. We recorded 17 Easy trials (26{,}882 frames) and 9 Hard trials (12{,}373 frames), and all of them were completed successfully. The two levels differ in two ways: first, the trash objects start in a standing position in Easy, but are lying flat or tilted in Hard (see Figure \ref{fig:tasks}); second, the trash bin is smaller in Hard. These changes make it harder to control the arm precisely. 

\subsubsection{Task Duration and Rhythm}

\begin{table}[h]
\centering
\caption{Task Duration: Easy vs.\ Hard}
\label{tab:duration}
\begin{tabular}{lcc}
\hline
Metric & Easy & Hard \\
\hline
Mean duration      & $40.2 \pm 8.7$\,s & $78.6 \pm 8.8$\,s \\
Min / Max          & 28.4\,s / 58.1\,s   & 66.5\,s / 95.2\,s \\
\hline
\end{tabular}
\end{table}

The mean task duration for the Hard group (78.6 s) is about twice as long as that of the Easy group (40.2 s). The time ranges for the two groups do not overlap at all: the longest Easy trial (58.1 s) is shorter than the shortest Hard trial (66.5 s), leaving a clear gap of roughly 8 seconds. This shows that the difficulty setting does a good job of capturing how complex the task is — harder tasks simply take more time to finish. The two groups also have similar standard deviations (Easy 8.7 s, Hard 8.8 s), which means that performance was equally steady in both cases. This supports the idea that the longer times are due to the task itself being harder, not because the operators were more or less consistent.

\subsubsection{Arm IK Activity Rate}

\begin{table}[h]
\centering
\caption{Arm IK Activity Rate: Easy vs.\ Hard}
\label{tab:activity}
\begin{tabular}{lcc}
\hline
Action type & Easy & Hard \\
\hline
Arm IK & 72.5\% & 82.5\% \\
\hline
\end{tabular}
\end{table}

The arm IK active frame ratio is substantially higher under Hard difficulty (82.5\% vs.\ 72.5\%, $+$10.0 percentage points). When target objects are in more complex poses, subjects must continuously adjust the arm end-effector position for precise grasping, leading to higher arm control engagement.

\subsubsection{State--Action Space Coverage Comparison}

\begin{table}[h]
\centering
\caption{Coverage Comparison: Easy vs.\ Hard}
\label{tab:coverage_cmp}
\begin{tabular}{lcc}
\hline
Subspace & Easy & Hard \\
\hline
Hand joints (12-dim)   & 95\% & 96\% \\
Arm IK position (6-dim) & 98\% & 100\% \\
Chassis motion (2-dim) & 100\% & 100\% \\
Robot position (3-dim) & 88\% & 88\% \\
\hline
\end{tabular}
\end{table}

Overall coverage rates are similar, but Arm IK position coverage reaches 100\% under Hard, higher than the 98\% of Easy. This suggests that the Hard difficulty drives subjects to explore a broader arm workspace to handle differently posed target objects, which is beneficial from a data-diversity perspective.

\begin{figure}[H]
\centering
\includegraphics[width=0.48\textwidth,height=7cm,keepaspectratio]{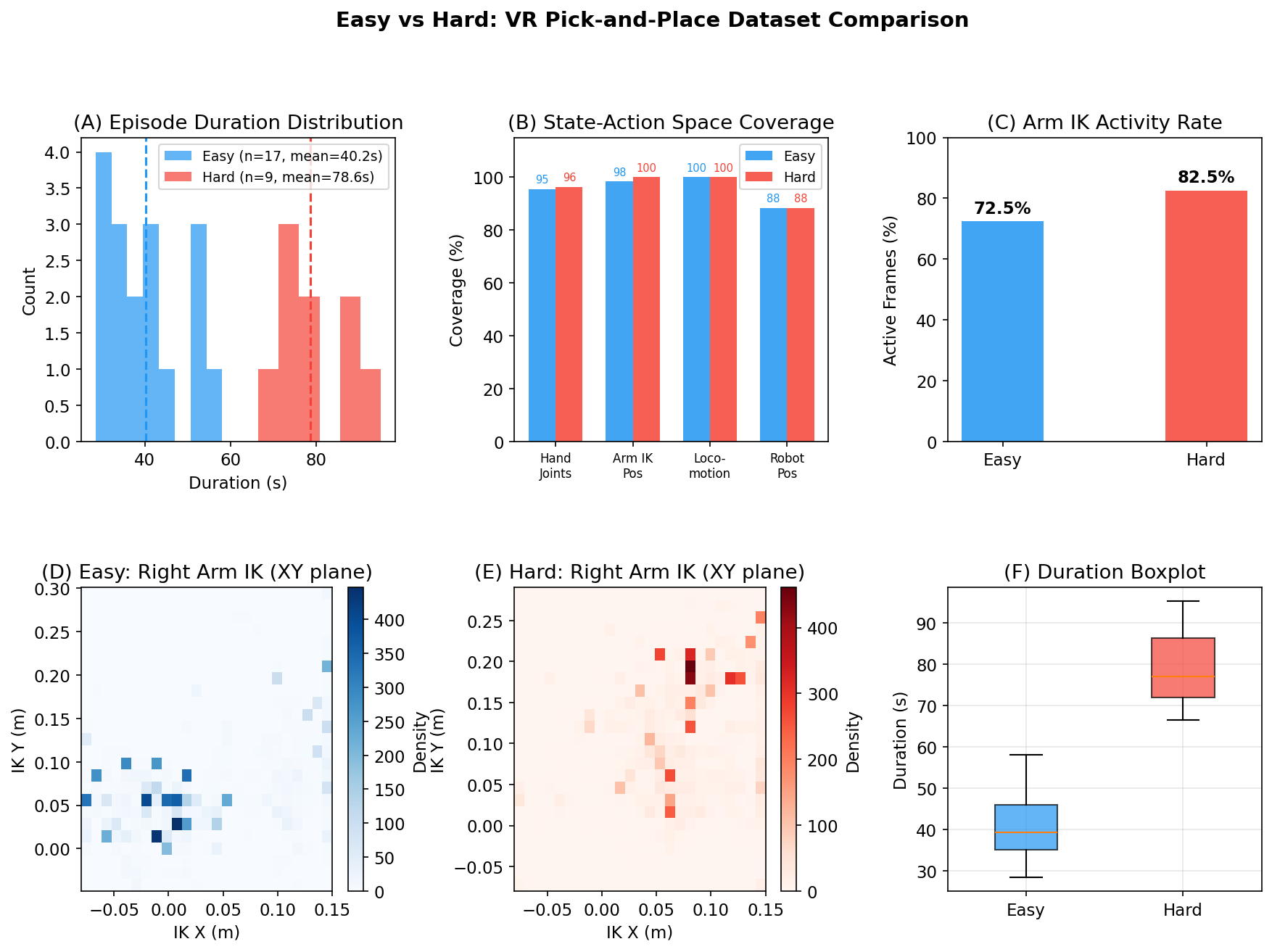}
\caption{Easy vs.\ Hard comparison: duration distribution, arm IK coverage, arm IK activity rate, and right-arm IK heatmaps.}
\label{fig:easy_vs_hard}
\end{figure}

These results demonstrate that the difficulty parameter has good behavioral discriminative power: as difficulty increases, task duration nearly doubles, arm control engagement rises significantly, and arm workspace coverage improves, exhibiting regular and interpretable behavioral changes. This validates the difficulty parameter as an effective axis for data stratification.

\subsection{Scene Randomness Analysis}

\subsubsection{Setup and Generation Mechanism}

This experiment uses the full dataset (17 Easy + 9 Hard, 26 episodes, 39{,}255 frames) to verify the controllability of procedural scene generation and the consistency of the collected data, analyzed from both mechanism and empirical perspectives.

At the mechanism level, scene randomness in each episode is governed by a set of bounded parameters. The room layout is determined by uniform sampling from a predefined template library; each template divides the room into functional zones (e.g., living room, kitchen), each with a preset set of allowed furniture categories and a maximum object count. Furniture is placed at random positions within the zone boundary and accepted only after passing three validity filters: downward ray-casting for height determination, overlap detection, and boundary checking. This guarantees physically feasible placements in every episode. The number of trash objects is sampled uniformly from a configurable range $[N_{\min}, N_{\max}]$ (default $[1, 5]$), object types are drawn from an allowed set, and positions are scattered uniformly within the central 50\% of the table surface. A difficulty parameter further controls the initial pose of trash objects (upright or lying flat) and the scale of the trash bin. All parameters are fixed before the experiment, and all random sampling is bounded, so results are fully reproducible under a fixed random seed.

\subsubsection{Empirical Distribution}

At the empirical level, Easy episode durations span 28.4\,s to 58.1\,s (mean $40.2 \pm 8.7$\,s) and Hard durations span 66.5\,s to 95.2\,s (mean $78.6 \pm 8.8$\,s); the combined robot root position spans a region of 5.8\,m $\times$ 6.0\,m in the XZ plane, indicating meaningful layout variation across episodes. Combined with the bin coverage rates above 88\% reported in Experiment~I, these observations confirm that the procedural generator introduces effective behavioral diversity within well-defined parameter bounds, and that the collected data are reliable for downstream use.

\section{CONCLUSIONS}
In this work, we proposed a gamified framework for embodied data collection that integrates procedural scene generation, VR-based interaction, automatic task evaluation, and structured data collection within a Unity environment. By transforming data collection into an interactive gameplay process, the framework reduces reliance on expert demonstrations and improves both scalability and accessibility. Experimental results demonstrate that the collected data achieves broad state-action coverage, that the difficulty parameter leads to measurable changes in human manipulation behavior, and that procedural generation introduces meaningful environmental diversity. These findings validate the feasibility of leveraging game mechanics for scalable embodied data acquisition.

In future work, we plan to expand task diversity and complexity, incorporate richer incentive and reward mechanisms to sustain long-term user participation, and extend the framework to multi-agent and more realistic interaction settings. Additionally, we will explore how the collected data can be utilized for downstream policy learning and sim-to-real transfer.

\addtolength{\textheight}{-12cm}   





\bibliographystyle{IEEEtran}
\bibliography{ref}

\end{document}